%% file: acl2020.tex
\documentclass[11pt,a4paper,usenames,dvipsnames]{article}
\usepackage[hyperref]{acl2020}
\usepackage{array}
\usepackage{amsmath}
\usepackage{booktabs}
\usepackage{framed}
\usepackage{graphicx}
\usepackage{latexsym}
\usepackage{makecell}
\usepackage{multirow}
\usepackage{subcaption}
\usepackage{times}
\usepackage{enumitem}
\usepackage{xspace}

\usepackage{soul}
\usepackage{microtype}
\usepackage{scalerel}

\newcommand\dataName{AmbER\xspace}
\newcommand\humanDataName{\dataName-\textit{H}\xspace}
\newcommand\nonhumanDataName{\dataName-\textit{N}\xspace}

\newif\ifcomments
\ifcomments
    \providecommand{\sameer}[2][]{{\protect\color{violet}{[sameer:\textbf{#1} #2]}}}
    \providecommand{\xiao}[2][]{{\protect\color{orange}{[xiao:\textbf{#1} #2]}}}
    \providecommand{\anthony}[2][]{{\protect\color{blue}{[anthony:\textbf{#1} #2]}}}
    \providecommand{\pallavi}[2][]{{\protect\color{teal}{[pallavi:\textbf{#1} #2]}}}
\else
    \providecommand{\sameer}[2][]{}
    \providecommand{\xiao}[2][]{}
    \providecommand{\anthony}[2][]{}
    \providecommand{\pallavi}[2][]{}
\fi

\aclfinalcopy
\title{Evaluating Entity Disambiguation and the Role of Popularity\\in Retrieval-Based NLP}

\newcommand{\authorspace}{\hspace{0.3cm}}

\author{
  \bf Anthony Chen\thanks{~~Work started during an internship at Apple.}$\:\,^{\includegraphics[scale=.06, clip]{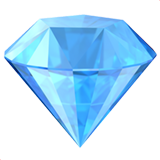}}$\authorspace{}
  \bf Pallavi Gudipati$^{\includegraphics[scale=.06, clip]{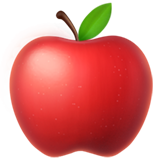}}$ \authorspace{}
  \bf Shayne Longpre$^{\includegraphics[scale=.06, clip]{images/Apple.png}}$
  \\
  \bf Xiao Ling$^{\includegraphics[scale=.06, clip]{images/Apple.png}}$ \authorspace
  \bf Sameer Singh$^{\includegraphics[scale=.06, clip]{images/Gem.png}}$  \vspace{2mm}
\\
  $^{\includegraphics[scale=.06, clip]{images/Gem.png}}$University of California, Irvine \authorspace{}
  $^{\includegraphics[scale=.06, clip]{images/Apple.png}}$Apple\\
  \{\href{mailto:anthony.chen@uci.edu}{\tt anthony.chen},
    \href{mailto:sameer@uci.edu}{\tt sameer}\}\href{mailto:anthony.chen@uci.edu}{\tt @uci.edu}\\
  \{\href{mailto:pgudipati@apple.com}{\tt pgudipati},
    \href{mailto:slongpre@apple.com}{\tt slongpre},
    \href{mailto:xiaoling@apple.com}{\tt xiaoling}\}\href{mailto:pgudipati@apple.com}{\tt @apple.com}
}

\begin{document}
\maketitle

\input{sections/00-abstract}
\input{sections/01-introduction}
\input{sections/02-motivation}

\input{sections/03-dataset}
\input{sections/04-experimental_setup}
\input{sections/05-results}
\input{sections/06-related_work}
\input{sections/conclusion}

\section*{Acknowledgements}
We would like to thank Jo Daiber, Michael Tu, Russ Webb, Matt Gardner, Robert Logan, Sherry Tongshuang Wu, and the anonymous reviewers for providing valuable feedback for our work.
This work is funded in part by the DARPA MCS program under Contract No. N660011924033 with the United States Office Of Naval Research.


\bibliography{rebiber-acl2020}
\bibliographystyle{acl_natbib}
\clearpage
\input{sections/appendix.tex}
\end{document}

%% file: sections/00-abstract.tex

\begin{abstract}
    Retrieval is a core component for open-domain NLP tasks.
    In open-domain tasks, multiple entities can share a name, making disambiguation an inherent yet under-explored problem.
    We propose an evaluation benchmark for assessing the entity disambiguation capabilities of these retrievers, which we call \textit{\textbf{Amb}iguous \textbf{E}ntity \textbf{R}etrieval (\dataName) sets}.
    We define an \dataName set as a collection of entities that share a name along with queries about those entities.
    By covering the set of entities for polysemous names, \dataName sets act as a challenging test of entity disambiguation.
    We create \dataName sets for three popular open-domain tasks: fact checking, slot filling, and question answering, and evaluate a diverse set of retrievers. 
    We find that the retrievers exhibit popularity bias, significantly under-performing on rarer entities that share a name, e.g., they
    are twice as likely to retrieve erroneous documents on queries for the less popular entity under the same name.
    These experiments on \dataName sets show their utility as an evaluation tool and highlight the weaknesses of popular retrieval systems.\footnote{The \dataName sets used in this paper and the code to generate them are available at \url{https://github.com/anthonywchen/AmbER-Sets}.}
\end{abstract}

%% file: sections/01-introduction.tex
\begin{figure}[t!]
    \small
    \begin{framed}
        \textbf{Q:} Which battle did \textcolor{red}{\textbf{Abe Lincoln}} fight in?\\
        \textbf{A:} World War II \\ 
        \textbf{Wikipedia Documents Ranked by BLINK:} \\
        \textbf{\texttt{1. Abraham\_Lincoln}} \\
        \texttt{2. Abraham\_Lincoln\_in\_the\_Black\_Hawk\_War} \\
        \texttt{3. Abraham\_Lincoln\_(captain)} \\
        \texttt{4. Benjamin\_Lincoln} \\
        \texttt{5. Lincoln\_Nebraska} \\
        \texttt{6. Lincoln\_England} \\\
        
        \textbf{Q:} What musical instrument does \textcolor{blue}{\textbf{Abe Lincoln}} play? \\
        \textbf{A:} Trombone \\ 
        \textbf{Wikipedia Documents Ranked by BLINK:} \\
        \texttt{1. Abraham\_Lincoln} \\
        \texttt{2. John\_Wilkes\_Booth} \\
        \texttt{3. Abe\_(musical)} \\
        \texttt{4. Nebraska} \\
        \texttt{5. Lincoln\_Nebraska} \\
        \textbf{\texttt{6. Abe\_Lincoln\_(musician)}}
    \end{framed}
    \vskip -3mm
    \caption{
        Queries for two entities (\textcolor{red}{president} \& \textcolor{blue}{musician}) with the name ``Abe Lincoln''.
        Retrieving the \textbf{gold document} involves disambiguating which ``Abe Lincoln'' each query is asking about.
        BLINK performs sub-optimally on the second query, as it ranks the document of the president over the gold document.
    }
    \vskip -3mm
    \label{fig:introduction:amber_example}
\end{figure}  

\section{Introduction}
    Substantial progress in NLP has been made on ``closed'' tasks, where queries are paired with relevant documents \citep{Rajpurkar2016SQuAD10, Dua2019DROPAR}.
    However, there is growing interest in ``open-domain'' tasks, where relevant documents need to be retrieved from a knowledge source before an NLP system can perform reasoning and produce an answer~\citep{Chen2017ReadingWT,Petroni2020KILTAB}.
    The open-domain setting better reflects real-world usage for tasks where relevant information is generally not provided (\textit{e.g.,} fact checking).
    Because success hinges on finding relevant documents, open-domain progress has been closely tied to improvements in retrieval systems\footnote{For example, replacing the BM25 retriever with DPR on Natural Questions increases exact match by 15 points.} \citep{Lee2019LatentRF,Karpukhin2020DensePR,Lewis2020RetrievalAugmentedGF}.

    A crucial challenge when interacting with a large knowledge source (\textit{e.g.}, Wikipedia) is entity ambiguity, the phenomenon where a single name can map to multiple entities.
    Resolving this ambiguity is referred to as entity disambiguation and is an important step for effective retrieval.
    For example, given the query \textit{``What musical  instrument does Abe Lincoln play?''}, documents about the musician should rank higher than other entities with the same name (Figure \ref{fig:introduction:amber_example}).
    Although entity disambiguation has been extensively studied in entity linking \citep{Hoffart2011RobustDO, Rao2013EntityLF, Sevgili2020NeuralEL} and search \citep{Balog2010OverviewOT, Balog2011OverviewOT}, in the context of open-domain NLP, it is unclear how good retrieval systems are when faced with queries with ambiguous entities.
    Evaluating entity ambiguity is challenging because the popularity of entities follows a long-tail (Figure \ref{fig:entity_disambiguation:popularity_plot}) and rare entities are seldom covered in naturally-occurring datasets.
    
    In this paper we introduce \dataName sets, a benchmark for evaluating the entity disambiguation capabilities of retrievers across multiple NLP tasks.
    Each \dataName set is a collection of Wikidata entities that share a name, and their corresponding queries for specific NLP tasks.
    For each set, we define the \textcolor{red}{\textbf{head}} entity as the most popular entity and \textcolor{blue}{\textbf{tail}} entities as the less popular ones. 
    By creating queries for multiple entities that share a name, \dataName sets provide an accurate test of entity disambiguation capabilities of retrievers and help assess the role of entity popularity in disambiguation.
    We show examples of \dataName sets for the question answering task in Table ~\ref{table:dataset:examples}.
    We automatically create \dataName sets by mining the Wikidata knowledge graph \citep{Vrandecic2014WikidataAF} for relevant names and entities, and leveraging task-specific templates to generate inputs for three tasks: fact checking, slot filling, and question answering (Figure \ref{fig:dataset:pipeline}).
    In total, our \dataName sets contain 80k task-specific queries which we align to the Wikipedia snapshot from KILT \citep{Petroni2020KILTAB}.

    We use \dataName sets to conduct a systematic study of various retrieval systems that operate under different principles, such as token overlap and dense embedding similarity.
    Retrievers perform very differently on \dataName sets in terms of absolute retrieval numbers, with Bootleg \citep{Orr2020BootlegCT}, an entity-linking-based retriever, performing best.
    Despite these differences, all retrievers exhibit a large degree of popularity bias, under-performing on inputs concerning tail entities.
    TF-IDF, a token-based retriever, performs about four times worse on tail entity inputs compared to head entity inputs.
    Even with Bootleg, the best performing retriever, performance on tail entities is still 1.5 times lower than on head entities.
    Our results on \dataName sets demonstrate that there is significant work to be done on making retrievers robust in handling entity disambiguation.

%% file: sections/02-motivation.tex
\begin{figure}[t!]
    \centering
  	\includegraphics[width=0.48\textwidth]{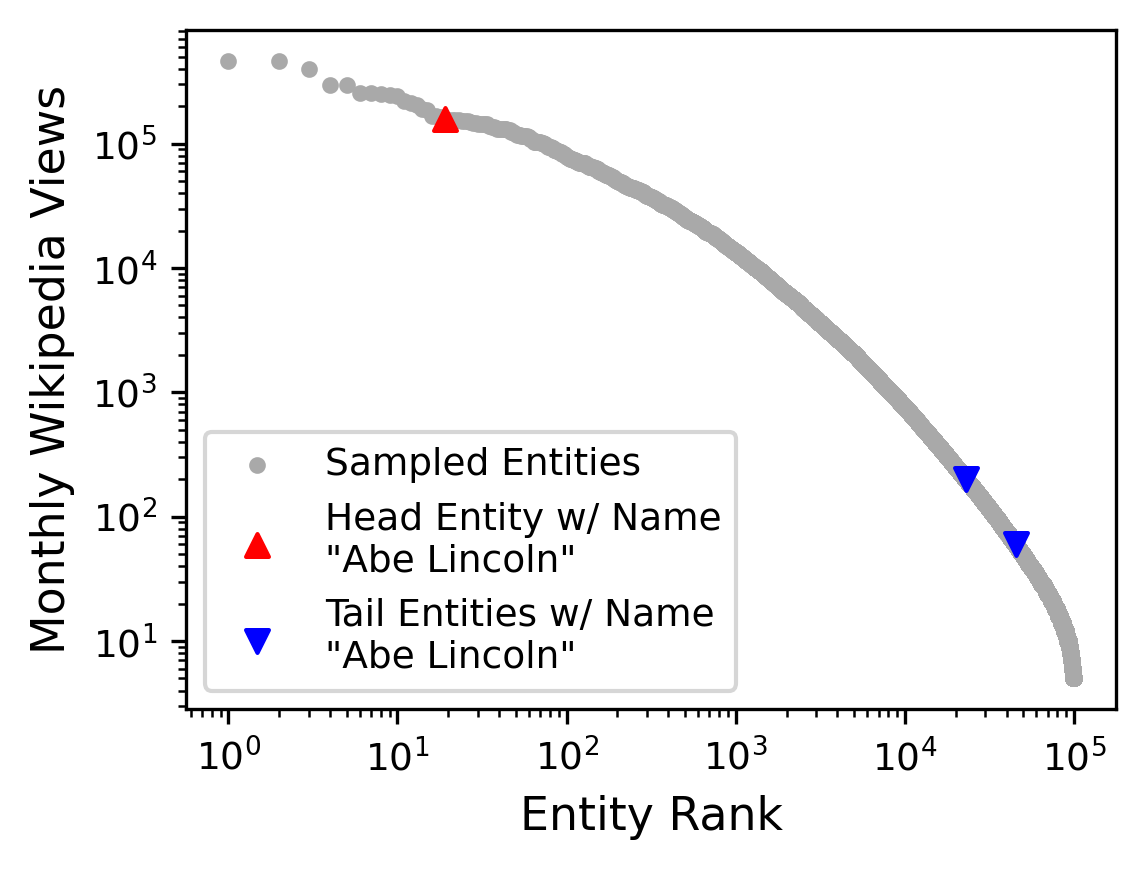}
    \vskip -3mm
    \caption{
        \textbf{The Long Tail of Entity Popularity:} Graph of the Wikipedia pageviews (in October 2019) for each Wikidata entity, ranked by popularity.
    	Gray are 100k randomly sampled entities, while red/blue are entities with the name ``Abe Lincoln''.
    }
    \vskip -3mm
    \label{fig:entity_disambiguation:popularity_plot}
\end{figure} 

\begin{table*}[t]
    \small
    \centering
    \begin{tabular}{lllll}
        \toprule 
         & \bf QID & \bf Input &  \bf Answer & \bf Gold Document \\
        \midrule
        \multirow{6}{*}{\bf \humanDataName} & \textcolor{red}{Q517} & What wars did \textbf{Napoleon} participate in? & Napoleon Wars & Napoleon\\
            & \textcolor{blue}{Q3335909} & What sport does \textbf{Napoleon} play? & Rugby & Napolioni\_Nalaga \\
            & \textcolor{blue}{Q3335909} & Which team does \textbf{Napoleon} play for? & Fiji National & Napolioni\_Nalaga \\
            \addlinespace
            
            & \textcolor{red}{Q117012} & What movement did \textbf{Yoko Ono} participate in? & Fluxus & Yoko\_Ono \\
            & \textcolor{blue}{Q16264827} & Which sport does \textbf{Yoko Ono} participate in? & Judo & Yoko\_Ono\_(judoka)  \\
        \midrule
        \multirow{7}{*}{\bf \nonhumanDataName} & \textcolor{red}{Q312} & Which industry is \textbf{Apple} in? & Electronics & Apple\_Inc. \\
            & \textcolor{blue}{Q532100} & What is the record label of \textbf{Apple}? & Page One & Apple\_(band)\\
            & \textcolor{blue}{Q7714007} & Who acted in \textbf{Apple}? & Ray Shell & The\_Apple\_(1980\_film) \\
            \addlinespace
            
            & \textcolor{red}{Q788822} & Who is a cast member on \textbf{Her}? & Steve Zissis & Her\_(film) \\
            & \textcolor{red}{Q788822} & Who is \textbf{Her}'s screenwriter? & Spike Jonze & Her\_(film) \\
            & \textcolor{blue}{Q28441308} & Who performed \textbf{Her}? & Aaron Tippin & Her\_(song) \\
        \bottomrule
    \end{tabular}
    \caption{
        \textbf{Examples of QA \dataName sets}.
        An \dataName set is a collection of entities that share a name, with instantiated queries for each entity.
        In this work, we use Wikidata to collect entities (QID).
        We also create queries for two more tasks, fact checking and slot filling (omitted from this table).
    }
    \label{table:dataset:examples}
\end{table*}

\section{\dataName Sets}
    Retrieving relevant documents from large knowledge sources such as Wikipedia is an important first step in the open-domain pipeline.
    An inherent problem in working with such sources is entity disambiguation: resolving a name (mention) to an entity in the knowledge source.
    Entity disambiguation can be challenging because many entities share a name, and the popularity of entities follows a long-tail distribution (Figure \ref{fig:entity_disambiguation:popularity_plot}).
    Despite the importance of entity disambiguation, it remains an understudied problem for open-domain NLP.
    We introduce \dataName sets for evaluating entity disambiguation capabilities of retrievers and analyze the role of entity popularity in disambiguation.
    
	\subsection{What is an \dataName Set?}
        We first provide an intuition for an \dataName set before concretely defining one.
        Consider two entities, a president and a musician, both of which have the name ``Abe Lincoln'' (Figure \ref{fig:introduction:amber_example}).
        Now, consider the query ``Which battle did Abe Lincoln fight in?'' and assume a retriever correctly returns the article about the president for this query.
        Simply because the correct document was retrieved does not mean a retriever has the ability to disambiguate between the president and the musician, as the president is much more popular.
        We should only be confident in its ability to disambiguate entities if we \textit{also} pose a query about the less popular musician and the retriever again returns the correct document (as opposed to the document about the president).

        Based on this intuition, we define an \dataName set as a collection of queries that satisfy the following:

		\begin{itemize}[leftmargin=10pt,topsep=0mm,itemsep=0mm]
			\item{
				\textbf{Criteria 1: Polysemous Name}:
				The queries in an \dataName set are all about entities that share a common name (\textit{e.g.}, Abe Lincoln).
			}
			\item{
			    \textbf{Criteria 2: Disparity in Popularity}:
			    An \dataName set contains queries about both the most popular entity for a name (the \textcolor{red}{head} entity), \textit{e.g.}, the president, and the less popular entities (the \textcolor{blue}{{tail}} entities), \textit{e.g.}, the musician.
			}
			\item{
				\textbf{Criteria 3: Resolvable Ambiguity}:
				The content of the query should be sufficient to resolve to the correct entity.
				The query \textit{``Which battle did Abe Lincoln fight in?''} satisfies this criteria, because there is only one Abe Lincoln that fought in a war, while \textit{``Where was Abe Lincoln born?''} does not since it applies to all Abe Lincolns.
			}
		\end{itemize}
		We provide examples of \dataName sets for the task of question answering in Table \ref{table:dataset:examples}.
        
    \subsection{Open-Domain Tasks}
        In this work, we create \dataName sets for three tasks: fact checking, slot filling, and question answering (Table \ref{table:baselines:training_data}).
		We consider these three tasks for three reasons.
		First, these three set of tasks are diverse in nature.
		In this work, slot filling is a generation task, question answering is a span selection task, and fact checking is a classification task.
		Second, the training sets available for each task are quite disparate.
		The largest fact checking training set, FEVER \citep{Thorne2018FEVERAL}, has 80k instances, while the slot filling dataset, T-REx \citep{ElSahar2018TRExAL}, has over 2 million instances.
		The final reason we study these three tasks is that their inputs are short and easy to create.
		
			\begin{table}[t!]
	    \small
	    \centering
        \begin{tabular}{l l l}
	        \toprule 
	        \bf Task & \bf Input Instance & \bf Output \\
	        \midrule
	            FC & John Mayer plays music. & True\\
	            SF & Nike [SEP] country & USA\\
	            QA & Whose face is on \$100 bill? & Benjamin Franklin\\
	        \bottomrule
        \end{tabular}
        \vskip -3mm
        \caption{Examples for each open-domain NLP task.}
	    \label{table:baselines:training_data}
    \vskip -3mm
	\end{table}

%% file: sections/03-dataset.tex
\begin{figure*}[t!h]
    \centering
  	\includegraphics[width=\textwidth]{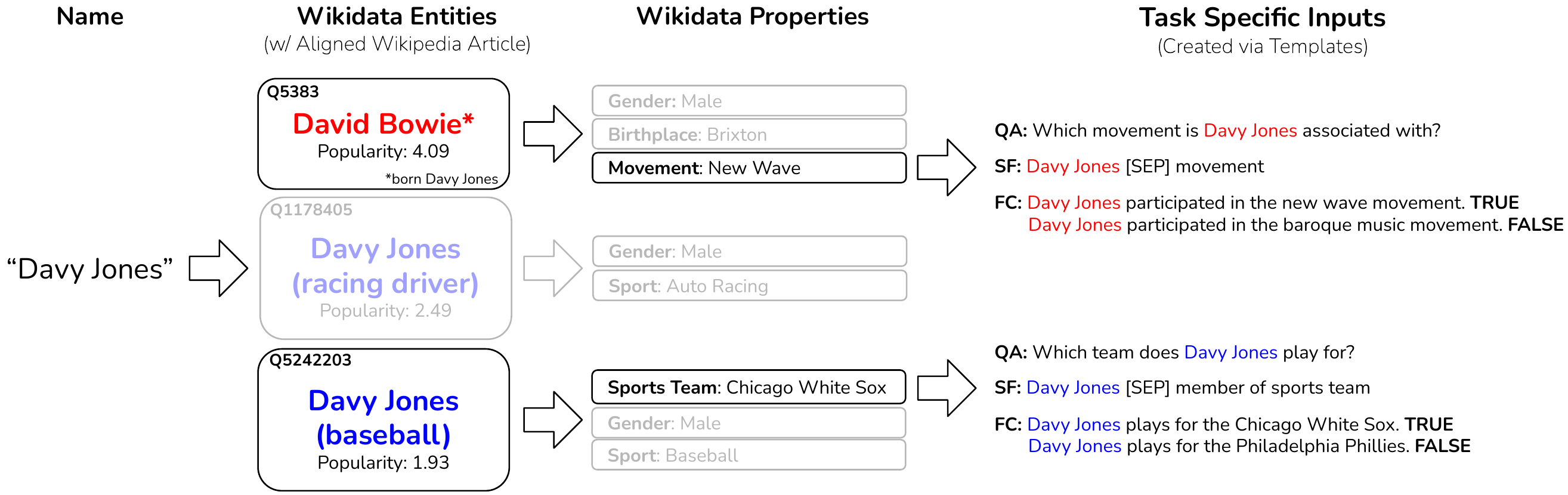}

    \caption{
    	\textbf{Automated creation of \dataName sets for three tasks.}
		We collect sets of entities from Wikipedia that share a name, where the most popular entity is the head entity (in red) and others are tail entities (in blue), 
		along with their properties and associated values. 
		We filter out properties that do not help distinguish entities in the set (gray-ed out), and remove entities that do not have any properties remaining. 
		From the remaining properties, we instantiate queries via templates for three tasks: question answering (QA), slot filling (SF), and fact checking (FC).
    }
            \vskip -2mm
    \label{fig:dataset:pipeline}
\end{figure*}

\section{Creating \dataName Sets}
    While \dataName sets can be manually created, doing so can be time-consuming, requiring a human to manually scour a knowledge base for polysemous names and related entities before manually writing queries for those entities.
    Instead, we present a pipeline for automatically creating \dataName sets using the Wikidata knowledge graph \citep{Vrandecic2014WikidataAF}.
    In this section, we describe two different \textit{collections} of \dataName sets, and discuss our automatic pipeline for creating \dataName sets.
    
	\subsection{Two Collections of \dataName Sets}
	    A natural question is ``How do retrievers handle entity ambiguity when two entities have the same entity type as opposed when they have different types?''.
	    To answer this question, we create two \textit{collections} of \dataName sets.
	    The first is \humanDataName, a collection of \dataName sets where all entities are humans.
	    The choice to restrict \humanDataName to humans is motivated by the fact that humans have properties that help distinguish themselves from other humans, generally based on occupation.
	    The second is \nonhumanDataName, a collection of \dataName sets where all entities contained are non-humans, and disambiguation of a name is between non-human entities with different entity types.
	    This is because a non-human entity, like a movie, does not generally have a single distinguishing property to distinguish from other movies.
	    This makes it natural to compare non-human entities to other non-human entities with different types.
	    We specify the entity types in each collection in Table \ref{table:dataset:PIDs}.
	    
	\subsection{Automatic Creation of \dataName Sets}
		\label{subsec:amber_creation}
	    We now describe a pipeline to automatically create \dataName sets for three tasks: fact checking, slot filling, and question answering.
	    We provide a visualization of the pipeline in Figure \ref{fig:dataset:pipeline}.

		\paragraph{Collecting Names and Entities}
    		We begin by collecting all entity aliases\footnote{Aliases are all possible names for an entity.} in Wikidata.
    		From these aliases, we filter for those that are shared by multiple Wikidata entities.
    		Each entity in Wikidata is represented by a unique QID.
    		The entities must have an entity type from Table~\ref{table:dataset:PIDs} depending on the collection we are collecting \dataName sets for.
            Each alias and associated entities form the basis for an \dataName set.
            Within each set, we define the head and tail entities based on the number of Wikipedia page views for the month of October 2019.
            We filter out \dataName sets where the percentage gap in popularity between the head entity and the most popular tail entity is less than 10\% to account for noise in the monthly page views.

        \begin{table}[t!]
            \small
            \centering
                \begin{tabular}{lllr}
                \toprule 
                & \textbf{Entity Type} & \textbf{Property} \textit{(PID)} & \textbf{Percent} \\
                \midrule
        	        \multirow{10}{*}{\bf\rotatebox[origin=c]{90}{\humanDataName}} & \multirow{10}{*}{Human} & instrument \textit{\small(P1303)} & 17.01 \\
        	        	& & movement \textit{\small(P135)} & 2.04 \\
        	        	& & appears in \textit{\small(P1441)} & 0.08 \\
        	        	& & killed by \textit{\small(P157)} & 0.19 \\
        	        	& & PhD student \textit{\small(P185)} & 0.42 \\
        				& & military branch \textit{\small(P241)} & 12.22 \\
        				& & sports position \textit{\small(P413)} & 12.82 \\
        				& & sports team \textit{\small(P54)} & 17.25 \\
        				& & battles or wars \textit{\small(P607)} & 12.29 \\
        				& & sport \textit{\small(P641)} & 25.68 \\
                \midrule
                     \multirow{18}{*}{\bf\rotatebox[origin=c]{90}{\nonhumanDataName}} & \multirow{3}{*}{Album} & performer \textit{\small(P175)} & 16.57 \\
                        & & record label \textit{\small(P264)} & 7.11 \\
                        & & tracklist \textit{\small(P658)} & 0.21 \\
            	        \addlinespace[1mm]
                        & \multirow{1}{*}{Business} & industry \textit{\small(P452)} & 0.65 \\
                        \addlinespace[1mm]
                        & \multirow{1}{*}{City} & population \textit{\small(P1082)} & 0.24 \\
                        \addlinespace[1mm]
                        & \multirow{2}{*}{Film} & cast member \textit{\small(P161)} & 27.14 \\
                        & & screenwriter \textit{\small(P58)} & 18.28 \\
                        \addlinespace[1mm]
                        & \multirow{1}{*}{Literary Work} & author \textit{\small(P50)} & 11.13 \\
                        \addlinespace[1mm]
                        & \multirow{1}{*}{Musical Group} & record label \textit{\small(P264)} & 2.1 \\
                        \addlinespace[1mm]
                        & \multirow{2}{*}{Song} & performer \textit{\small(P175)} & 4.42 \\
                        & & record label \textit{\small(P264)} & 0.62 \\
                        \addlinespace[1mm]
                        & \multirow{3}{*}{TV Series} & cast member \textit{\small(P161)} & 2.01 \\
                        & & \# seasons \textit{\small(P2437)} & 1.85 \\
                        & & screenwriter \textit{\small(P58)} & 0.21 \\
                        \addlinespace[1mm]
                        & \multirow{1}{*}{Written Work} & author \textit{\small(P50)} & 7.43 \\
                \bottomrule
                \end{tabular}
            \caption{
                \textbf{Distinguishing Properties} selected to create queries based on whether they are sufficient to resolve ambiguity.
                We provide the percent breakdown of how often each property occurs in each \dataName collection.
            }
            \vskip -4mm
        	\label{table:dataset:PIDs}
        \end{table}

        \paragraph{Collecting Distinguishing Properties}
			We gather properties and associated values for each entity from Wikidata.
			We only retain properties that are in a specified list (Table \ref{table:dataset:PIDs}), as they are useful for resolving ambiguity \textit{(Criteria 3)}.
			We also filter a property if two entities within an \dataName set have that property, ensuring that the remaining properties can be used to disambiguate between entities with the same name.
			These properties are used to instantiate the queries.
	
		\paragraph{Aligning Entities to Wikipedia}
		    We use the KILT Wikipedia snapshot \citep{Petroni2020KILTAB} as the knowledge source for \dataName sets for better reproducibility.
		    Each Wikipedia document in KILT has an associated QID.
			For each entity, we find all Wikipedia documents with that associated QID.
			After this alignment, we apply a round of filtering on the tuples.
			For each tuple, we check that the value of the tuple is within the first 350 tokens of the aligned Wikipedia article.
			If not, we remove the tuple.\footnote{This reduces the number of tuples for \humanDataName from 17,079 to 5,942 and for \nonhumanDataName from 22,219 to 13,804.}
			Aligned Wikipedia articles that contain the tuple value serve as gold documents.
	
		\paragraph{Instantiating \dataName Instances}
		    Recall that our goal was to create \dataName sets for three tasks: fact checking, slot filling, and question answering.
		    We are able to create queries for all three tasks simultaneously using the collected Wikidata tuples.
			For question answering and fact checking, we use templates based on properties to instantiate inputs.
			Three of the authors wrote a template each for each property for the two tasks.
			Duplicate templates are removed, resulting in an average of 3 question answering templates per property and 2.7 fact checking templates per property.
			See Appendix \ref{appendex:templates} for the complete list of templates.
		    
			For slot filling, we create a single input from each Wikidata tuple by concatenating the \dataName set name with the property name, and using the value of the tuple as the answer.
			For question answering, we also create a single input for each tuple by filling in the template with the \dataName set name and using the value of the tuple as the answer.
			For fact checking, we create two inputs for each tuple, one claim that is true using the tuple value and one claim that is false.
			The false claim is created by finding the most popular value for the tuple property that does not match the tuple value\footnote{
			    The most popular instrument in Wikidata is piano. 
			    Therefore, given the true claim \textit{``Abe Lincoln played the trombone.''}, the false claim would be \textit{``Abe Lincoln played the piano.''}.
			}.

	\begin{table}[t]
	    \small
	    \centering
	    \begin{tabular}{l@{\hspace{0\tabcolsep}}rr}
	        \toprule
	        & \bf \humanDataName & \bf \nonhumanDataName\\
	        \midrule
	        \# \dataName Sets & 2,093 & 5,237 \\
	        \addlinespace[2mm]
	        Averages per \dataName Set \\
	        \ldots \# entities & 2.98 & 2.42 \\
	        \ldots \# entities w/ properties & 2.03 & 2.06 \\
	        \ldots \# properties & 2.84 & 2.64 \\

	        \addlinespace[2mm]
	        \# Input Queries & 23,768 & 55,216 \\
			\ldots Question Answering (QA) & 5,942 & 13,804 \\
	        \ldots Slot Filling (SF) & 5,942 & 13,804 \\
	        \ldots Fact checking (FC) & 11,884 & 27,608\\
	        \bottomrule
	    \end{tabular}
	    \caption{Statistics of \dataName collections.}
	    \label{table:dataset:statistics}
            \vskip -3mm
	\end{table}

	\subsection{Dataset Statistics}
		We provide statistics for \dataName sets in 
		Table~\ref{table:dataset:statistics}.
		On average, each \dataName set has about three entities that share the same name. 
		Of these three entities, on average, only two have properties after filtering.
		In total, our \dataName sets contain about 80k task-specific input queries.

    \subsection{Limitations}
        Since our pipeline is automated and relies on Wikipedia and Wikidata, there are a few limitations worth noting.
        \dataName sets will be affected by incompleteness of the knowledge source, sometimes resulting ambiguous queries if a property is missing from Wikidata, but answerable from Wikipedia text. 
        For this reason, we only select a few properties for each type (Table~\ref{table:dataset:PIDs}).
        Second, even though we author multiple templates for each property, the reliance on these templates limits the syntactic diversity in the queries (not a critical concern, since we are only evaluating existing models).
        Also, we use Wikipedia page views as a proxy for real-world popularity of entities.
        Defining popularity in this way may be problematic, as page views for an entity can fluctuate, and may make our pipeline difficult to generalize to other knowledge sources, where this information may not be available.
        
        Several design choices in creating \dataName sets are worth further investigation. 
        We limit \dataName sets to a pre-specified list of entity types and properties to ensure that entities in an \dataName set are distinguishable. 
        This precludes other properties that may be useful in distinguishing entities, reducing the diversity in \dataName sets.
        Another design choice is we allow any alias in Wikidata to form an \dataName sets, however, not all aliases are canonical ways to refer to the entity.
        For instance, Shaquille O'Neal has the unusual alias ``The Big Cactus'', potentially leading to a somewhat unrealistic query \textit{``What sport did The Big Cactus play?''}.
        We plan to revisit the these design choices in future work.

%% file: sections/04-experimental_setup.tex
\begin{table*}[t!]
    \small
    \centering
        \begin{tabular}{l l rrrr rrrr rrrr} 
            \toprule 
            \multirow{2}{*}{\bf Collection}  & \multirow{2}{*}{\bf  Retriever} & \multicolumn{4}{c}{\bf Fact Checking (FC)} & \multicolumn{4}{c}{\bf Slot Filling (SF)} & \multicolumn{4}{c}{\bf Question Answering (QA)} \\
            \cmidrule(lr){3-6}
            \cmidrule(lr){7-10}
            \cmidrule(lr){11-14}
            & & All &  \textcolor{red}{Head} &  \textcolor{blue}{Tail} & $\forall$ & All &  \textcolor{red}{Head} &  \textcolor{blue}{Tail} & $\forall$ & All &  \textcolor{red}{Head} &  \textcolor{blue}{Tail} & $\forall$  \\
            \midrule
            \multirow{4}{*}{\bf  \humanDataName} & TF-IDF & 17.3 &  28.5 &  8.2 & 0.0 & 18.8 &  31.9 &  8.1 & 0.0 & 16.7 &  28.2 &  7.3 & 0.1\\
            & DPR    & 18.1 &  23.9 &  13.3 & 0.1 & 8.0 &  11.6 &  5.1 &  0.3 & 13.1 &  19.6 &  7.9 & 1.1 \\
            & BLINK  & \bf 55.9 & \bf 64.4 & \bf 49.0 & \bf 5.6 & 38.2 & 57.0 &  22.9 & 11.5 & 31.7 & 40.5 & 24.6 &  6.6\\
            & Bootleg  & 34.8 & 43.0 & 28.2 & 0.7 & \textbf{56.5} & \textbf{63.9 }& \textbf{50.6} &  \textbf{25.3} & \textbf{67.2} & \textbf{77.1} & \textbf{59.1} & \textbf{36.1} \\
            \addlinespace
            \multirow{4}{*}{\bf  \nonhumanDataName} & TF-IDF & 9.4 &  13.6  &  4.9 & 0.0 &  13.4 &  21.0 &  5.2 & 0.2 &  13.9 &  21.7 &  5.4 & 0.3 \\
            & DPR    & \bf 36.9 & \bf 48.0 & \bf 24.8 & \bf 4.4 & 29.9 & 40.9 & 18.0 & 6.0 & 36.2 & 49.2 &  22.2 & 9.3 \\
            & BLINK  & 11.7 &  13.9 &  9.4 &  0.0 & 5.7 &  7.3 &  3.9 & 0.7 & 35.2 &  44.7 & 24.9 & 10.1 \\
            & Bootleg  & 3.5 & 4.6 & 2.4 & 0.0 & \bf 52.3 & \bf 61.3 & \bf 42.5 & \bf 22.4 & \bf 59.8 & \bf 69.5 & \bf 49.3 & \bf     29.0  \\
            \bottomrule
        \end{tabular}
    \caption{
        \textbf{Top-1 retrieval results} on each collection of \dataName sets. 
        We report accuracy@1 results on all instances as well as results on instances about head entities and instances about tail entities. 
        We also report a set-level metric, \textit{all correct} ($\forall$), the percentage of \dataName sets where \emph{all} inputs had the correct document retrieved.}
            \vskip -3mm
    \label{table:results:acc_at_1}
\end{table*}    

\section{Evaluation Setup}
	\paragraph{Retrieval Systems} 
    	The primary focus of this work is to evaluate entity ambiguity of retrieval systems. 
    	We consider four retrievers based on different retrieval paradigms.
    	The first three are TF-IDF, a token-based retriever using sparse embeddings, DPR \citep{Karpukhin2020DensePR}, a dense embedding based retriever, and BLINK \citep{Wu2020ZeroshotEL}, a linker-based retriever which ranks documents based on input entities.
    	These three retrievers have been thoroughly evaluated on a number of open-domain tasks in \citet{Petroni2020KILTAB} with no obvious winner across tasks.
    	Encouraged by the disambiguation success on rare entities by \citet{Orr2020BootlegCT}, we also evaluate a retriever based on Bootleg, another entity linker.
    	We provide additional details about these retrievers in Appendix \ref{appendix:retriever}.
	
	\paragraph{Downstream Models}
        The dominant approach to open-domain tasks is a two-stage process where a retriever first finds relevant documents, 
        followed by a downstream model that processes these documents to produce an answer.
    	We evaluate the end-to-end performance on \dataName sets by training downstream NLP models on our tasks of interest.
    	For fact checking, we fine-tune a BERT classifier ~\citep{Devlin2019BERTPO} on FEVER \citep{Thorne2018FEVERAL}.
    	For question answering, we fine-tune a RoBERTa model \citep{Liu2019RoBERTaAR} on Natural Questions \citep{Kwiatkowski2019NaturalQA}.
    	For slot filling, a generation task, we fine-tune a BART model~\citep{Lewis2020BARTDS} on T-Rex \citep{ElSahar2018TRExAL}.
    	We provide example training instances in Table \ref{table:baselines:training_data} and additional details on the models in Appendix \ref{appendix:downstream_model}.
    	We use the AllenNLP and HuggingFace Transformers library to finetune our downstream models~\citep{Gardner2018AllenNLPAD, wolf-etal-2020-transformers}.

%% file: sections/05-results.tex
\begin{table}[t]
    \small
    \centering
        \begin{tabular}{l l c@{\hspace{0.75\tabcolsep}}c c@{\hspace{0.75\tabcolsep}}c c@{\hspace{0.75\tabcolsep}}c}
        \toprule 
        & & \multicolumn{2}{c}{\bf FC} & \multicolumn{2}{c}{\bf SF} & \multicolumn{2}{c}{\bf QA} \\
        & &  \textcolor{red}{Head} & \textcolor{blue}{Tail} & \textcolor{red}{Head} & \textcolor{blue}{Tail} & \textcolor{red}{Head} & \textcolor{blue}{Tail} \\
        \midrule
    	\multirow{4}{*}{\textit{\textbf{H\**}}} & TF-IDF & 19.5 & 67.5 & 28.2 & 75.7 & 27.9 & 76.1 \\
    	& DPR   & 1.2 & 10.0 & 2.3 & 23.8 & 2.6 & 27.0 \\
    	& BLINK & 9.8 & 32.2 & 14.0 & 58.2 & 4.4 & 27.6 \\
    	& Bootleg & 6.2 & 24.7& 9.3& 30.5& 3.7& 28.7\\
    	\midrule 
    	\multirow{4}{*}{\textit{\textbf{N\**}}} & TF-IDF & 10.1 & 49.9 & 22.0 & 76.9 & 23.0 & 76.8 \\
    	& DPR   & 6.2 & 32.2 & 9.1 & 48.3 & 8.7 & 44.0 \\
    	& BLINK & 5.8 & 22.8 & 5.1 & 32.2 & 5.5 & 31.9 \\   
    	& Bootleg & 7.7 & 26.1& 16.1& 36.2& 7.8& 31.6\\
        \bottomrule
        \end{tabular}
    \**~\textbf{\textit{H}} represents \humanDataName and \textbf{\textit{N}} represents \nonhumanDataName.
    \caption{
        \textbf{Entity confusion} measures the \% of queries the gold document ranks \textbf{worse} (lower) than a document for another entity with the same name (\textit{i.e.}, another entity in the \dataName set).
        Retrievers are four times as likely to exhibit this when dealing tail queries.
    }
            \vskip -3mm
    \label{table:results:entity_interference}
\end{table} 

\begin{figure}[t!]
    \centering
  	\includegraphics[width=0.48\textwidth]{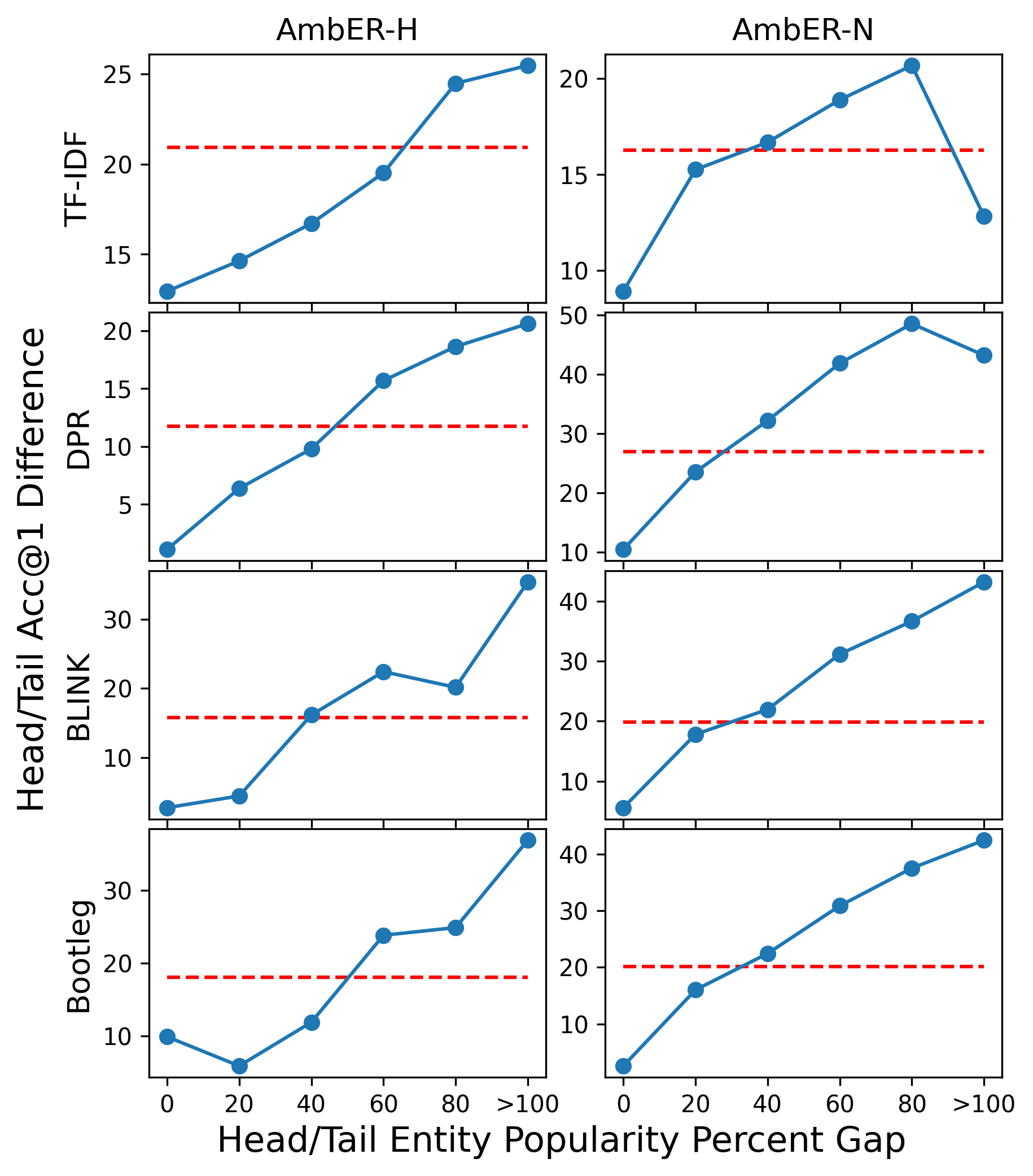}
    \caption{
        \textbf{Popularity Gap vs Retrieval Gap.}
         We bin QA queries of pairs of head and tail entities based on the popularity gap between the entities. 
         For each bin, we calculate the retrieval accuracy@1 difference on the head and tail queries.
         Larger popularity gaps tend to lead to a wider gaps in retrieval performance.
         The red line is retrievers' performance gaps between head and tail queries on the entire collection.
    }
    \label{fig:results:popularity_gap_results}
\end{figure}

\begin{table}[tb]
    \small
    \centering
        \begin{tabular}{l l lrrr}
        \toprule 
        & \multirow{2}{*}{\textbf{Task}} & \multirow{2}{*}{\bf System} & \multicolumn{3}{c}{\bf Results} \\
        & & & All &  \textcolor{red}{Head} &  \textcolor{blue}{Tail} \\
        \midrule
        \multirow{7}{*}{\textbf{\textit{H}}} & \multirow{2}{*}{FC} & BERT (Oracle) & 77.7 &  73.6 &  80.3 \\
        & & BERT + BLINK & 59.8 &  60.1 &  57.7 \\
        \addlinespace[1mm]
        & \multirow{2}{*}{SF} & BART (Oracle) & 83.9 &  85.0 &  83.5 \\
        & & BART + BLINK & 34.4 &  38.2 &  32.6 \\
        \addlinespace[1mm]
        & \multirow{2}{*}{QA}& BERT (Oracle) & 71.4 &  77.7 &  83.0 \\
        & & BERT + BLINK & 27.5 &  33.8 &  22.3\\
        \midrule
        \multirow{7}{*}{\textbf{\textit{N}}} & \multirow{2}{*}{FC} & BERT (Oracle) & 66.6 &  63.9 &  69.5 \\
        & & BERT + DPR & 60.9 &  61.4 & 60.4 \\
        \addlinespace[1mm]
        & \multirow{2}{*}{SF} & BART (Oracle) & 82.1 &  80.1 &  84.3 \\
        & & BART + DPR & 18.6 & 18.6  & 18.6  \\
        \addlinespace[1mm]
        & \multirow{2}{*}{QA} & BERT (Oracle) & 83.5 &  85.1 &  81.8 \\
        &  & BERT + DPR & 26.0 &  31.3 &  20.4\\
        \bottomrule
        \end{tabular}
    \caption{\textbf{End-to-end performance on \dataName sets}. We evaluate systems in an oracle setting, where the gold document is provided, and a retrieval setting, where 20 documents are provided from a retriever.}
    \label{table:results:downstream_explicit}
\end{table}    

\section{Results}
    In this section, we evaluate existing open-domain NLP pipelines using \dataName sets.
    We also conduct a user study to evaluate the quality of the queries in the \dataName sets.

	\paragraph{Top Document Retrieval}
		We report retrieval performance in Table \ref{table:results:acc_at_1} in terms of retriever accuracy@1 (the \% of instances where the first retrieved document is the gold document).
		For each task, we report values on the entire \dataName set (``All''), as well as instances corresponding only to ``Head'' entities or to ``Tail'' entities. 
		We also report a metric we call \textit{all correct} ($\forall$), the fraction of \dataName sets in which all queries had the correct document retrieved.
		All retrievers do better on head entities compared to tail entities.
		Since BLINK, Bootleg, and DPR are initialized using pre-trained language models, they may have a predisposition towards being biased to more popular entities.
		However, we find TF-IDF also does better on head entities, perhaps because more popular entities have longer Wikipedia pages, possibly increasing term-frequency scores.
		Second, there are large discrepancies between a retriever's performance on different tasks for an \dataName collection.
		For instance, DPR does substantially worse on slot filling compared to its performance on question answering.
		This is surprising  since queries for all tasks are created from the same set of Wikidata tuples.
		Finally, we find that retrievers are mostly incorrect on getting all the queries in a set correct, with some receiving a $\forall$ score of 0 on some tasks.
		Overall, we find that the Bootleg retriever on average does the best across tasks, however there is significant scope for improvement.

	\paragraph{Entity Confusion}
	    To explicitly evaluate whether retrievers get confused by entities in the same \dataName set, we compute \textit{entity confusion} for retrievers defined as the percentage of queries where the retriever ranks a document for an incorrect entity \emph{from the same \dataName set} over the gold document (Table \ref{table:results:entity_interference}). 
		We find that across retrievers, tasks, and  \dataName collections, entity confusion is twice as high for tail entity inputs. 
		This result indicates that the popularity of an entity for a given name plays a significant role in retrieval performance.
		
	\paragraph{Effect of Popularity Gap}
		Since the difference in popularity between the head and tail entities can vary considerably, these results obfuscate the effect of the \textit{size} of the popularity gap. 
		We explore how the gap in popularity between head and tail entities translates to the gaps in performance on their associated queries.
		For a head entity with popularity $p_h$ and a tail entity with popularity $p_t$ from the same \dataName set, we calculate popularity gap, $\frac{p_h - p_t}{p_t}$, and bin associated head/tail inputs based on the gap\footnote{Bin width of 20\%. Queries with a popularity gap higher than 100\% are binned into the highest bin.}.
		For each bin, we calculate the difference in accuracy@1 between the head and tail entity queries.
		Results for QA \dataName sets (Figure \ref{fig:results:popularity_gap_results}) show that there is a strong correlation between the popularity gap and the difference in performance. 
		
	\paragraph{End to End Results}	
		We evaluate end to end performance in several evaluation settings with all results provided in Table \ref{table:results:downstream_explicit}.
        The metrics used are F1 for slot filling and question answering and accuracy for fact checking.
		In the ``oracle'' setting, we directly provide the downstream NLP model the gold document, and find that the gap between head entities and tail entities is fairly small. 
		This suggests that in closed NLP settings, where the gold document is known, entity disambiguation is not a major concern. 

		In the regular retrieval setting, we provide the model the top 20 documents as ranked by a retrieval system (BLINK and DPR), and find that retrievers still perform better on head entity queries (see Appendix \ref{appendix:retrieval_at_20}).
        The downstream systems that use retrieved documents display a noticeable gap in end-to-end performance between head and tail entity inputs.
		This is expected, as retrieval systems perform worse on tail entities.
		
    \paragraph{User Study}
        \dataName sets are created in a largely automatic process, raising questions about data quality.
        To address these questions, we conduct a small user study on \dataName sets to evaluate whether the queries are resolvable by humans.
        We present a query from a QA \dataName set along with three documents for the entities from the same \dataName set, one of which is the gold document.
        We first ask the user to select the relevant document, then we ask the user to select an answer span from the selected document.
        In total, we asked 7 subjects to examine about 120 queries across \humanDataName and \nonhumanDataName, and computed their accuracy in selecting the correct document and answer (Table \ref{table:results:human_study}).
        We also compare retrievers for this task, \textit{i.e.} select from $3$ documents for the same queries, and find that humans perform very well on the document selection task compared to retrievers on both sets.
        We also compare the accuracy of answer selection, and see that the closed domain NLP model (fine-tuned BERT) is as almost accurate as humans on the same set of queries\footnote{The relatively low answer score is due to artifacts in using EM for QA evaluation, and is consistent with human performance on span selection~\cite{Rajpurkar2016SQuAD10}).}.
        This further confirms that closed NLP models are not the source of bias towards head entities, but the retrievers are.

\newcolumntype{P}[1]{>{\centering\arraybackslash}p{#1}}
\begin{table}[t!]
    \small
    \centering
        \begin{tabular}{l cc cc}
        \toprule
        \multirow{2}{*}{\bf System} & \multicolumn{2}{c}{\bf \humanDataName} & \multicolumn{2}{c}{\bf \nonhumanDataName} \\
            & Doc Acc. & EM & Doc Acc. & EM\\
        \midrule
            TF-IDF & 43.3    & - & 50.3 & -\\
            DPR & 69.1 & - & 68.3 & - \\
            BLINK & 69.1 & - & 74.1 & -\\
            Bootleg & 79.6 & - & 73.1 & -\\    
            \addlinespace
            BERT & - & 71.8 & - & 75.5\\
            \addlinespace
            \bf Human & \bf 100 & \bf 78.8 & \bf 97.9 & \bf 77.5 \\
        \bottomrule
        \end{tabular}
    \caption{\textbf{User study on \dataName QA.} Humans are nearly perfect in identifying the correct document for each query (Doc Acc), while existing retrievers frequently fail. When the gold document is provided to downstream NLP models (BERT), they do almost as well as humans in answering the question (EM).}
	\label{table:results:human_study}
\end{table}

%% file: sections/06-related_work.tex

\section{Related Work}
	
	\paragraph{Entity Ambiguity} 
	As previously mentioned, entity ambiguity is when a single name can match multiple entities in a knowledge source.
	Entity ambiguity has been most studied in the context of entity linking \citep{Rao2013EntityLF}.
	To improve disambiguation, entity linkers have included auxiliary information such as entity types \citep{Onoe2020FineGrainedET} and entity descriptions \citep{Logeswaran2019ZeroShotEL}.
	A recent thread of work aims to study how language models recall and leverage information about names and entities.
	\citet{Prabhakaran2019PerturbationSA} shows that names can have a measurable effect on the prediction of sentiment analysis systems.
	\citet{Shwartz2020YouAG} demonstrates that pre-trained language models implicitly resolve entity ambiguity by grounding names to entities based on the pre-training corpus.
	The problem of entity ambiguity also appears implicitly in entity-centric tasks such as determining the semantic relatedness between entities \citep{Hoffart2012KOREKO} and entity-oriented search \citep{Balog2010OverviewOT, Balog2011OverviewOT}.
	We draw inspiration from these works by studying entity ambiguity in the context of open-domain NLP.
	
	\paragraph{Popularity Bias} 
	System's that perform worse on the long-tail suffer from what is known as popularity bias. 
	This problem has been studied extensively in the recommendation systems literature, where recommendation systems are known to often ignore the long-tail of products and instead recommend very popular items \citep{Abdollahpouri2017ControllingPB, Chen2020BiasAD}.
	This has the effect of unfairly hurting users who would prefer these less-popular items \citep{Abdollahpouri2019TheUO, Nematzadeh2018HowAP}.
	We explore popularity bias from the angle of retrieval as opposed to recommendation, and find popularity bias exists in retrieval systems.
	
	\paragraph{Open-Domain Ambiguity} Ambiguity is an inherent problem when it comes to open-domain reasoning.
	\citet{Min2020AmbigQAAA} showed that half of instances sampled from Natural Questions are ambiguous, with multiple correct answers.
	\dataName sets are similar in that the ambiguity is in terms of the entity in the query, however, in contrast to Natural Questions, \dataName set inputs have been constructed such that the ambiguity is resolvable.

	\paragraph{Challenge Sets} There have been many evaluation sets specifically designed to assess a model's ability to handle a specific phenomenon \citep{Naik2018StressTE, Zhao2018GenderBI, McCoy2019RightFT, Warstadt2019BLiMPTB, Richardson2020ProbingNL, Jeretic2020AreNL, ribeiro-etal-2019-red}.
	Some of these challenge sets, similar to \dataName sets, use templates to generate a large amount of evaluation data quickly \citep{Richardson2020ProbingNL, McCoy2019RightFT, ribeiro-etal-2020-beyond}.
	\dataName sets can be viewed as a challenge set for assessing open-domain systems' ability to handle entity ambiguity.
	

%% file: sections/conclusion.tex

\section{Conclusion}
    Entity ambiguity is an inherent problem in retrieval, as many entities can share a name.
    For evaluating disambiguation capabilities of retrievers, we introduce \dataName sets; an \dataName set is a collection of task-specific queries about entities that share a name, but the queries have sufficient content to resolve the correct entity.  
    We create a broad range of \dataName sets, covering many entity types, with input queries for three open-domain NLP tasks: fact checking, slot filling, and question answering.
    Our experiments demonstrate the struggles of current retrievers in handling entity ambiguity.
    In particular, we find that the popularity of an entity in relation to other entities that share a name plays a significant role during disambiguation.
    For instance, we find that all tested retrievers are about twice as likely to retrieve erroneous documents when dealing with less popular entities than the most popular entity with the same name.
    Future goals include improving entity disambiguation capabilities of retrievers, perhaps more directly incorporating ideas from entity linking and coreference resolution.
    The \dataName sets and the code for the generation pipeline is available at \url{https://github.com/anthonywchen/AmbER-Sets}.

%% file: sections/appendix.tex
\section*{Appendix}
\appendix

\section{Top-20 Retrieval Results}
\label{appendix:retrieval_at_20}
We provide results for top-20 retrieval in Table \ref{table:appendix:retrieval_at_20}.
Top-20 retrieval is used for providing documents in the end-to-end evaluation setting.
In this setting, retrieval accuracy measures whether a gold document appears in one of the top-20 retrieved documents.
Similar to top-1 retrieval, retrievers continue to perform better on head queries.

\section{Task Specific Templates}
\label{appendex:templates}
Table \ref{table:appendix:templates} contains the templates used to instantiate the task-specific inputs.
Templates were written on a per-property basis.
We note that many of the properties share templates that are very similar.

\section{Computational Resources}
    All experiments (\textit{e.g.}, training baselines, generating \dataName sets, etc.) were conducted on a machine with 500 GB of RAM,  64 CPUs, and using an NVIDIA TitanRTX with 24 GB of RAM. 
    Retrieval on a collection of \dataName sets takes about 12 hours for the most time-consuming retriever, BLINK.
    Training a downstream model takes roughly 5 hours and inference on a collection of \dataName sets takes less than 30 minutes.

\begin{table*}[tbh]
    \small
    \centering
        \begin{tabular}{l l c@{\hspace{\tabcolsep}}c@{\hspace{\tabcolsep}}c@{\hspace{\tabcolsep}}c  c@{\hspace{\tabcolsep}}c@{\hspace{\tabcolsep}}c@{\hspace{\tabcolsep}}c  c@{\hspace{\tabcolsep}}c@{\hspace{\tabcolsep}}c@{\hspace{\tabcolsep}}c}
            \toprule 
            \multirow{2}{*}{\bf Collection}  & \multirow{2}{*}{\bf  Retriever} & \multicolumn{4}{c}{\bf Fact Checking} & \multicolumn{4}{c}{\bf Slot Filling} & \multicolumn{4}{c}{\bf Question Answering} \\
            \cmidrule(lr){3-6}
            \cmidrule(lr){7-10}
            \cmidrule(lr){11-14}
            & & All &  \textcolor{red}{Head} &  \textcolor{blue}{Tail} & $\forall$ & All &  \textcolor{red}{Head} &  \textcolor{blue}{Tail} & $\forall$ & All &  \textcolor{red}{Head} &  \textcolor{blue}{Tail} & $\forall$  \\
            \midrule
            \multirow{3}{*}{\bf  \humanDataName} & TF-IDF & 65.8 &  78.5 & 55.4 & 26.7  & 72.0 & 83.5  & 62.5  & 55.6 & 72.6  & 82.0 & 64.8 & 55.9 \\
            & DPR    & 39.8 & 51.0 & 30.6 & 4.1 & 26.6 & 37.0 & 18.1 & 6.8 & 36.1  & 49.3 & 25.3 & 9.6\\
            & BLINK  & 78.6 & 82.0 & 76.0 & 43.8 & 73.3 & 73.9 & 72.8 & 64.6 & 58.8  & 60.3 & 57.5 & 32.2 \\
            & Bootleg  &  \textbf{96.5} & \textbf{97.6}  & \textbf{95.6} & \textbf{93.2} & \textbf{96.6} & \textbf{97.7} & \textbf{95.7 }&\textbf{ 93.6 }& \textbf{96.5} & \textbf{97.6} & \textbf{95.6} & \textbf{93.5} \\
            \midrule
            \multirow{3}{*}{\bf  \nonhumanDataName} & TF-IDF & 50.8 & 57.0 & 44.1 & 12.0 & 46.8 & 53.4 & 39.7  & 35.3 & 52.0 & 59.1  & 44.4 & 40.7 \\
            & DPR       & 62.3 & 75.8 & 47.7 & 27.8 & 57.3 & 71.4 & 42.0 & 29.4  & 63.4 & 77.9 &  47.8 & 37.2 \\
            & BLINK     & 33.5 & 38.7 & 27.9 & 1.3 & 18.2 & 21.5 & 14.6 & 5.8 & 74.7 & 80.6 & 68.3 & 53.0 \\
            & Bootleg   & \textbf{79.3} & \textbf{80.2 } & \textbf{78.4} & \textbf{61.5} & \textbf{89.6} & \textbf{91.9} & \textbf{87.1}  &\textbf{ 85.3}  & \textbf{83.8} & \textbf{83.6} & \textbf{84.1} & \textbf{71.1}\\
            \bottomrule
        \end{tabular}
    \caption{Top-20 retrieval results measuring retrieval accuracy and $\forall$.}
    \label{table:appendix:retrieval_at_20}
\end{table*}

\section{Retriever Details}
\label{appendix:retriever}
    For BLINK, DPR, and TF-IDF, we use the retriever code in the KILT repository released by Facebook\footnote{\url{https://github.com/facebookresearch/KILT}}. 
    For Bootleg, we use the code provided by the Hazy Research group\footnote{\url{https://github.com/HazyResearch/bootleg}}.

\section{Downstream Model Details}
    \label{appendix:downstream_model}
    For question answering, we train a RoBERTa-Large model on Natural Questions.
    We use the negative documents in Natural Questions to train a ``no-answer'' classifier using the \texttt{[CLS]} token.
    During inference, we take the highest-scoring span where the answer is not classified as ``no-answer''.
    For slot filling, we train a BART-base model.
    For each slot filling instance, we train with the top non-gold document retrieved by TF-IDF as a negative document.
    For this negative document, we train the model to generate a ``none'' token, and during inference, we take the highest scoring answer that is not ``none''.
    For fact checking, we train a three-way (\textit{i.e.}, \texttt{SUPPORTS}, \texttt{REFUTES}, \texttt{NEUTRAL}) BERT-base classifier. 
    Similar to slot filling, we train with the top non-gold document retrieved by TF-IDF as a negative document and train the model to classify this negative document as  \texttt{NEUTRAL}.
    During inference, we take the highest scoring prediction that is not  \texttt{NEUTRAL}.
    When training baselines models, we do not tune over hyperparameters and train with a batch size of 32 for 3 epochs.

\begin{table*}[tb]
    \small
    \centering
    \begin{tabular}{llp{2.3in}p{2.3in}@{\hspace{0.75\tabcolsep}}}
        \toprule 
        & \bf Property & \bf Question Answering Template & \bf Fact Checking Template \\
        \midrule
        \multirow{36}{*}{\bf \rotatebox[origin=c]{90}{\humanDataName}} & instrument & Which musical instrument did \textit{\textbf{\$name}} play? \newline What musical instrument does \textit{\textbf{\$name}} play? \newline What instrument does \textit{\textbf{\$name}} play? & \textit{\textbf{\$name}} plays the \textit{\textbf{\$object}}. \newline \textit{\textbf{\$name}} plays the musical instrument \textit{\textbf{\$object}}. \newline The \textit{\textbf{\$object}} is played by \textit{\textbf{\$name}}. \\\addlinespace[2mm]
            & movement & What movement did \textit{\textbf{\$name}} participate in? \newline Which movement is \textit{\textbf{\$name}} associated with? \newline What movement is \textit{\textbf{\$name}} associated with? & \textit{\textbf{\$name}} was a member of the \textit{\textbf{\$object}} movement. \newline \textit{\textbf{\$name}} participated in the \textit{\textbf{\$object}} movement. \newline \textit{\textbf{\$name}} was a part of the \textit{\textbf{\$object}} movement. \\\addlinespace[2mm]
            & appears in & What works does the fictional entity \$name appear in? \newline What work is the character \$name present in? \newline Which work was the character \$name in? & \$name is a character in \$object. \newline \$name is a fictional character in \$object. \newline \$object features the fictional character \$name. \\\addlinespace[2mm]
            & doctoral student & Who were the doctoral students of \textit{\textbf{\$name}}? \newline Who are \textit{\textbf{\$name}}'s doctoral students? \newline Who did \textit{\textbf{\$name}} advise? & \textit{\textbf{\$name}} has a doctoral student named \textit{\textbf{\$object}}. \newline \textit{\textbf{\$name}}'s doctoral student is \textit{\textbf{\$object}}. \newline \textit{\textbf{\$name}} advised their student \textit{\textbf{\$object}}. \\\addlinespace[2mm]
            & military branch & What branch of the military does \textit{\textbf{\$name}} belong to? \newline Which military branch does \textit{\textbf{\$name}} belong to? \newline What military branch is \textit{\textbf{\$name}} affiliated with? & \textit{\textbf{\$name}} is a member of the \textit{\textbf{\$object}}. \newline \textit{\textbf{\$name}} belongs to the military branch \textit{\textbf{\$object}}. \newline \textit{\textbf{\$name}} belongs to the \textit{\textbf{\$object}} branch of the military. \\\addlinespace[2mm]
            & sports position & What is the position that \textit{\textbf{\$name}} plays? \newline What position does \textit{\textbf{\$name}} play? \newline Which position does \textit{\textbf{\$name}} play? & \textit{\textbf{\$name}} plays the \textit{\textbf{\$object}} position. \newline \textit{\textbf{\$name}} plays as a \textit{\textbf{\$object}}. \\\addlinespace[2mm]
            & sports team & \textit{\textbf{\$name}} plays for which team? \newline What team does \textit{\textbf{\$name}} play for? \newline Which team does \textit{\textbf{\$name}} play for? & \textit{\textbf{\$name}} is a player on the \textit{\textbf{\$object}}. \newline \textit{\textbf{\$name}} plays for the \textit{\textbf{\$object}} team. \newline \textit{\textbf{\$name}} plays for the \textit{\textbf{\$object}}. \\\addlinespace[2mm]
            & battles or wars & What were the wars that \textit{\textbf{\$name}} participated in? \newline Which battle did \textit{\textbf{\$name}} fight in? \newline Which war did \textit{\textbf{\$name}} fight? & \textit{\textbf{\$name}} fought in the \textit{\textbf{\$object}}. \newline \textit{\textbf{\$name}} fought in \textit{\textbf{\$object}}. \\\addlinespace[2mm]
            & sport & Which sport does \textit{\textbf{\$name}} participate in? \newline Which sport does \textit{\textbf{\$name}} play? \newline What sport does \textit{\textbf{\$name}} play? & \textit{\textbf{\$name}} plays \textit{\textbf{\$object}}. \newline \textit{\textbf{\$name}} plays the sport \textit{\textbf{\$object}}. \\
        \midrule
        
        \multirow{30}{*}{\bf \rotatebox[origin=c]{90}{\nonhumanDataName}} & performer  & Who performs \textit{\textbf{\$name}}? \newline Who is the performer of \textit{\textbf{\$name}}? \newline Who performed \textit{\textbf{\$name}}? & \textit{\textbf{\$object}} performs in \textit{\textbf{\$name}}. \newline \textit{\textbf{\$object}} is the performer of \textit{\textbf{\$name}} . \newline \textit{\textbf{\$name}} was performed by \textit{\textbf{\$object}}. \\\addlinespace[2mm]
            & record label & What is the record label of \textit{\textbf{\$name}}.? \newline What is the record label for \textit{\textbf{\$name}}? \newline \textit{\textbf{\$name}} belongs to which record label? & \textit{\textbf{\$object}} is the record label for \textit{\textbf{\$name}}. \newline \textit{\textbf{\$name}}'s record label is \textit{\textbf{\$object}}. \\\addlinespace[2mm]
            & tracklist & What song appears in the album \textit{\textbf{\$name}}? \newline What song appears on \textit{\textbf{\$name}}? \newline What are the tracks in \textit{\textbf{\$name}}? & \textit{\textbf{\$name}} belongs to \textit{\textbf{\$object}} tracklist. \newline \textit{\textbf{\$object}} is on the release of \textit{\textbf{\$name}} . \newline \textit{\textbf{\$object}} is a song in the \textit{\textbf{\$name}} tracklist. \\\addlinespace[2mm]
            & industry & Which industry is \textit{\textbf{\$name}} in? \newline In what industry is \textit{\textbf{\$name}}? \newline What is \textit{\textbf{\$name}}'s industry? & \textit{\textbf{\$name}} is in the industry of \textit{\textbf{\$object}}. \newline The company \textit{\textbf{\$name}} is in the \textit{\textbf{\$object}} industry. \newline \textit{\textbf{\$name}}'s industry is \textit{\textbf{\$object}}. \\\addlinespace[2mm]
            & population & What is the total population of \textit{\textbf{\$name}}? \newline What is the population of \textit{\textbf{\$name}}? \newline How many people live in \textit{\textbf{\$name}}? & The population of \textit{\textbf{\$name}} is \textit{\textbf{\$object}}. \newline \textit{\textbf{\$name}}'s population is \textit{\textbf{\$object}}. \newline \textit{\textbf{\$name}} has a population of \textit{\textbf{\$object}}. \\\addlinespace[2mm]
            & cast member & Who acted in \textit{\textbf{\$name}}? \newline Who is a cast member on \textit{\textbf{\$name}}? \newline Who starred in \textit{\textbf{\$name}}? & \textit{\textbf{\$object}} was a cast member in \textit{\textbf{\$name}}. \newline \textit{\textbf{\$object}} appeared in \textit{\textbf{\$name}}. \newline \textit{\textbf{\$object}} acted in \textit{\textbf{\$name}}. \\\addlinespace[2mm]
            & screenwriter & Who was the screenwriter for \textit{\textbf{\$name}}? \newline Who was screenwriter for \textit{\textbf{\$name}}? \newline Who is \textit{\textbf{\$name}}'s screenwriter? & \textit{\textbf{\$name}}'s screenwriter is \textit{\textbf{\$object}}. \newline \textit{\textbf{\$object}} wrote the screenplay of \textit{\textbf{\$name}}. \newline \textit{\textbf{\$object}} screenwrote \textit{\textbf{\$name}}. \\\addlinespace[2mm]
            & \# seasons &How many seasons are there in \textit{\textbf{\$name}}? \newline How many seasons does \textit{\textbf{\$name}} have? \newline How many seasons were there in \textit{\textbf{\$name}}? & There were \textit{\textbf{\$object}} seasons in \textit{\textbf{\$name}}. \newline \textit{\textbf{\$name}} has \textit{\textbf{\$object}} seasons. \\\addlinespace[2mm]
            & author & Who is the author of \textit{\textbf{\$name}}? \newline Who wrote \textit{\textbf{\$name}}? \newline Who authored \textit{\textbf{\$name}}? & \textit{\textbf{\$name}} wrote \textit{\textbf{\$object}}. \newline \textit{\textbf{\$name}} is written by \textit{\textbf{\$object}}. \newline \textit{\textbf{\$object}} authored \textit{\textbf{\$name}}. \\
        \bottomrule
    \end{tabular}
    \caption{
        Templates used to instantiate the task-specific inputs.
    }
    \label{table:appendix:templates}
\end{table*}

%% file: acl2020.bbl
\begin{thebibliography}{41}
\expandafter\ifx\csname natexlab\endcsname\relax\def\natexlab#1{#1}\fi

\bibitem[{Abdollahpouri et~al.(2017)Abdollahpouri, Burke, and
  Mobasher}]{Abdollahpouri2017ControllingPB}
Himan Abdollahpouri, Robin Burke, and Bamshad Mobasher. 2017.
\newblock \href {https://doi.org/10.1145/3109859.3109912} {Controlling
  popularity bias in learning-to-rank recommendation}.
\newblock In \emph{Proceedings of the Eleventh {ACM} Conference on Recommender
  Systems, RecSys 2017, Como, Italy, August 27-31, 2017}, pages 42--46. {ACM}.

\bibitem[{Abdollahpouri et~al.(2019)Abdollahpouri, Mansoury, Burke, and
  Mobasher}]{Abdollahpouri2019TheUO}
Himan Abdollahpouri, Masoud Mansoury, Robin Burke, and Bamshad Mobasher. 2019.
\newblock \href {https://arxiv.org/abs/1907.13286} {The unfairness of
  popularity bias in recommendation}.
\newblock \emph{arXiv preprint arXiv:1907.13286}.

\bibitem[{Balog et~al.(2010)Balog, Serdyukov, and
  de~Vries}]{Balog2010OverviewOT}
K.~Balog, Pavel Serdyukov, and Arjen~P. de~Vries. 2010.
\newblock \href {https://trec.nist.gov/pubs/trec19/papers/ENTITY.OVERVIEW.pdf}
  {Overview of the trec 2010 entity track}.
\newblock In \emph{TREC}.

\bibitem[{Balog et~al.(2011)Balog, Serdyukov, and
  de~Vries}]{Balog2011OverviewOT}
K.~Balog, Pavel Serdyukov, and Arjen~P. de~Vries. 2011.
\newblock \href {https://trec.nist.gov/pubs/trec20/papers/ENTITY.OVERVIEW.pdf}
  {Overview of the trec 2011 entity track}.
\newblock In \emph{TREC}.

\bibitem[{Chen et~al.(2017)Chen, Fisch, Weston, and Bordes}]{Chen2017ReadingWT}
Danqi Chen, Adam Fisch, Jason Weston, and Antoine Bordes. 2017.
\newblock \href {https://doi.org/10.18653/v1/P17-1171} {Reading {W}ikipedia to
  answer open-domain questions}.
\newblock In \emph{Proceedings of the 55th Annual Meeting of the Association
  for Computational Linguistics (Volume 1: Long Papers)}, pages 1870--1879,
  Vancouver, Canada. Association for Computational Linguistics.

\bibitem[{Chen et~al.(2020)Chen, Dong, lei Wang, Feng, Wang, and
  He}]{Chen2020BiasAD}
J.~Chen, Hande Dong, Xiao lei Wang, Fuli Feng, Ming-Chieh Wang, and X.~He.
  2020.
\newblock \href {https://arxiv.org/abs/2010.03240} {Bias and debias in
  recommender system: A survey and future directions}.
\newblock \emph{arXiv preprint arXiv:2010.03240}.

\bibitem[{Ciampaglia et~al.(2018)Ciampaglia, Nematzadeh, Menczer, and
  Flammini}]{Nematzadeh2018HowAP}
Giovanni~Luca Ciampaglia, Azadeh Nematzadeh, Filippo Menczer, and Alessandro
  Flammini. 2018.
\newblock \href {https://www.nature.com/articles/s41598-018-34203-2} {How
  algorithmic popularity bias hinders or promotes quality}.
\newblock \emph{Scientific Reports}, 8.

\bibitem[{Devlin et~al.(2019)Devlin, Chang, Lee, and
  Toutanova}]{Devlin2019BERTPO}
Jacob Devlin, Ming-Wei Chang, Kenton Lee, and Kristina Toutanova. 2019.
\newblock \href {https://doi.org/10.18653/v1/N19-1423} {{BERT}: Pre-training of
  deep bidirectional transformers for language understanding}.
\newblock In \emph{Proceedings of the 2019 Conference of the North {A}merican
  Chapter of the Association for Computational Linguistics: Human Language
  Technologies, Volume 1 (Long and Short Papers)}, pages 4171--4186,
  Minneapolis, Minnesota. Association for Computational Linguistics.

\bibitem[{Dua et~al.(2019)Dua, Wang, Dasigi, Stanovsky, Singh, and
  Gardner}]{Dua2019DROPAR}
Dheeru Dua, Yizhong Wang, Pradeep Dasigi, Gabriel Stanovsky, Sameer Singh, and
  Matt Gardner. 2019.
\newblock \href {https://doi.org/10.18653/v1/N19-1246} {{DROP}: A reading
  comprehension benchmark requiring discrete reasoning over paragraphs}.
\newblock In \emph{Proceedings of the 2019 Conference of the North {A}merican
  Chapter of the Association for Computational Linguistics: Human Language
  Technologies, Volume 1 (Long and Short Papers)}, pages 2368--2378,
  Minneapolis, Minnesota. Association for Computational Linguistics.

\bibitem[{Elsahar et~al.(2018)Elsahar, Vougiouklis, Remaci, Gravier, Hare,
  Laforest, and Simperl}]{ElSahar2018TRExAL}
Hady Elsahar, Pavlos Vougiouklis, Arslen Remaci, Christophe Gravier, Jonathon
  Hare, Frederique Laforest, and Elena Simperl. 2018.
\newblock \href {https://www.aclweb.org/anthology/L18-1544} {{T}-{RE}x: A large
  scale alignment of natural language with knowledge base triples}.
\newblock In \emph{Proceedings of the Eleventh International Conference on
  Language Resources and Evaluation ({LREC} 2018)}, Miyazaki, Japan. European
  Language Resources Association (ELRA).

\bibitem[{Gardner et~al.(2018)Gardner, Grus, Neumann, Tafjord, Dasigi, Liu,
  Peters, Schmitz, and Zettlemoyer}]{Gardner2018AllenNLPAD}
Matt Gardner, Joel Grus, Mark Neumann, Oyvind Tafjord, Pradeep Dasigi,
  Nelson~F. Liu, Matthew Peters, Michael Schmitz, and Luke Zettlemoyer. 2018.
\newblock \href {https://doi.org/10.18653/v1/W18-2501} {{A}llen{NLP}: A deep
  semantic natural language processing platform}.
\newblock In \emph{Proceedings of Workshop for {NLP} Open Source Software
  ({NLP}-{OSS})}, pages 1--6, Melbourne, Australia. Association for
  Computational Linguistics.

\bibitem[{Hoffart et~al.(2012)Hoffart, Seufert, Nguyen, Theobald, and
  Weikum}]{Hoffart2012KOREKO}
Johannes Hoffart, Stephan Seufert, Dat~Ba Nguyen, Martin Theobald, and Gerhard
  Weikum. 2012.
\newblock \href {https://doi.org/10.1145/2396761.2396832} {{KORE:} keyphrase
  overlap relatedness for entity disambiguation}.
\newblock In \emph{21st {ACM} International Conference on Information and
  Knowledge Management, CIKM'12, Maui, HI, USA, October 29 - November 02,
  2012}, pages 545--554. {ACM}.

\bibitem[{Hoffart et~al.(2011)Hoffart, Yosef, Bordino, F{\"u}rstenau, Pinkal,
  Spaniol, Taneva, Thater, and Weikum}]{Hoffart2011RobustDO}
Johannes Hoffart, Mohamed~Amir Yosef, Ilaria Bordino, Hagen F{\"u}rstenau,
  Manfred Pinkal, Marc Spaniol, Bilyana Taneva, Stefan Thater, and Gerhard
  Weikum. 2011.
\newblock \href {https://www.aclweb.org/anthology/D11-1072} {Robust
  disambiguation of named entities in text}.
\newblock In \emph{Proceedings of the 2011 Conference on Empirical Methods in
  Natural Language Processing}, pages 782--792, Edinburgh, Scotland, UK.
  Association for Computational Linguistics.

\bibitem[{Jeretic et~al.(2020)Jeretic, Warstadt, Bhooshan, and
  Williams}]{Jeretic2020AreNL}
Paloma Jeretic, Alex Warstadt, Suvrat Bhooshan, and Adina Williams. 2020.
\newblock \href {https://doi.org/10.18653/v1/2020.acl-main.768} {Are natural
  language inference models {IMPPRESsive}? {L}earning {IMPlicature} and
  {PRESupposition}}.
\newblock In \emph{Proceedings of the 58th Annual Meeting of the Association
  for Computational Linguistics}, pages 8690--8705, Online. Association for
  Computational Linguistics.

\bibitem[{Karpukhin et~al.(2020)Karpukhin, Oguz, Min, Lewis, Wu, Edunov, Chen,
  and Yih}]{Karpukhin2020DensePR}
Vladimir Karpukhin, Barlas Oguz, Sewon Min, Patrick Lewis, Ledell Wu, Sergey
  Edunov, Danqi Chen, and Wen-tau Yih. 2020.
\newblock \href {https://doi.org/10.18653/v1/2020.emnlp-main.550} {Dense
  passage retrieval for open-domain question answering}.
\newblock In \emph{Proceedings of the 2020 Conference on Empirical Methods in
  Natural Language Processing (EMNLP)}, pages 6769--6781, Online. Association
  for Computational Linguistics.

\bibitem[{Kwiatkowski et~al.(2019)Kwiatkowski, Palomaki, Redfield, Collins,
  Parikh, Alberti, Epstein, Polosukhin, Devlin, Lee, Toutanova, Jones, Kelcey,
  Chang, Dai, Uszkoreit, Le, and Petrov}]{Kwiatkowski2019NaturalQA}
Tom Kwiatkowski, Jennimaria Palomaki, Olivia Redfield, Michael Collins, Ankur
  Parikh, Chris Alberti, Danielle Epstein, Illia Polosukhin, Jacob Devlin,
  Kenton Lee, Kristina Toutanova, Llion Jones, Matthew Kelcey, Ming-Wei Chang,
  Andrew~M. Dai, Jakob Uszkoreit, Quoc Le, and Slav Petrov. 2019.
\newblock \href {https://doi.org/10.1162/tacl_a_00276} {Natural questions: A
  benchmark for question answering research}.
\newblock \emph{Transactions of the Association for Computational Linguistics},
  7:452--466.

\bibitem[{Lee et~al.(2019)Lee, Chang, and Toutanova}]{Lee2019LatentRF}
Kenton Lee, Ming-Wei Chang, and Kristina Toutanova. 2019.
\newblock \href {https://doi.org/10.18653/v1/P19-1612} {Latent retrieval for
  weakly supervised open domain question answering}.
\newblock In \emph{Proceedings of the 57th Annual Meeting of the Association
  for Computational Linguistics}, pages 6086--6096, Florence, Italy.
  Association for Computational Linguistics.

\bibitem[{Lewis et~al.(2020{\natexlab{a}})Lewis, Liu, Goyal, Ghazvininejad,
  Mohamed, Levy, Stoyanov, and Zettlemoyer}]{Lewis2020BARTDS}
Mike Lewis, Yinhan Liu, Naman Goyal, Marjan Ghazvininejad, Abdelrahman Mohamed,
  Omer Levy, Veselin Stoyanov, and Luke Zettlemoyer. 2020{\natexlab{a}}.
\newblock \href {https://doi.org/10.18653/v1/2020.acl-main.703} {{BART}:
  Denoising sequence-to-sequence pre-training for natural language generation,
  translation, and comprehension}.
\newblock In \emph{Proceedings of the 58th Annual Meeting of the Association
  for Computational Linguistics}, pages 7871--7880, Online. Association for
  Computational Linguistics.

\bibitem[{Lewis et~al.(2020{\natexlab{b}})Lewis, Perez, Piktus, Petroni,
  Karpukhin, Goyal, K{\"{u}}ttler, Lewis, Yih, Rockt{\"{a}}schel, Riedel, and
  Kiela}]{Lewis2020RetrievalAugmentedGF}
Patrick S.~H. Lewis, Ethan Perez, Aleksandra Piktus, Fabio Petroni, Vladimir
  Karpukhin, Naman Goyal, Heinrich K{\"{u}}ttler, Mike Lewis, Wen{-}tau Yih,
  Tim Rockt{\"{a}}schel, Sebastian Riedel, and Douwe Kiela. 2020{\natexlab{b}}.
\newblock \href
  {https://proceedings.neurips.cc/paper/2020/hash/6b493230205f780e1bc26945df7481e5-Abstract.html}
  {Retrieval-augmented generation for knowledge-intensive {NLP} tasks}.
\newblock In \emph{Advances in Neural Information Processing Systems 33: Annual
  Conference on Neural Information Processing Systems 2020, NeurIPS 2020,
  December 6-12, 2020, virtual}.

\bibitem[{Liu et~al.(2019)Liu, Ott, Goyal, Du, Joshi, Chen, Levy, Lewis,
  Zettlemoyer, and Stoyanov}]{Liu2019RoBERTaAR}
Yinhan Liu, Myle Ott, Naman Goyal, Jingfei Du, Mandar Joshi, Danqi Chen, Omer
  Levy, Mike Lewis, Luke Zettlemoyer, and Veselin Stoyanov. 2019.
\newblock \href {https://arxiv.org/abs/1907.11692} {Roberta: A robustly
  optimized bert pretraining approach}.
\newblock \emph{arXiv preprint arXiv:1907.11692}.

\bibitem[{Logeswaran et~al.(2019)Logeswaran, Chang, Lee, Toutanova, Devlin, and
  Lee}]{Logeswaran2019ZeroShotEL}
Lajanugen Logeswaran, Ming-Wei Chang, Kenton Lee, Kristina Toutanova, Jacob
  Devlin, and Honglak Lee. 2019.
\newblock \href {https://doi.org/10.18653/v1/P19-1335} {Zero-shot entity
  linking by reading entity descriptions}.
\newblock In \emph{Proceedings of the 57th Annual Meeting of the Association
  for Computational Linguistics}, pages 3449--3460, Florence, Italy.
  Association for Computational Linguistics.

\bibitem[{McCoy et~al.(2019)McCoy, Pavlick, and Linzen}]{McCoy2019RightFT}
Tom McCoy, Ellie Pavlick, and Tal Linzen. 2019.
\newblock \href {https://doi.org/10.18653/v1/P19-1334} {Right for the wrong
  reasons: Diagnosing syntactic heuristics in natural language inference}.
\newblock In \emph{Proceedings of the 57th Annual Meeting of the Association
  for Computational Linguistics}, pages 3428--3448, Florence, Italy.
  Association for Computational Linguistics.

\bibitem[{Min et~al.(2020)Min, Michael, Hajishirzi, and
  Zettlemoyer}]{Min2020AmbigQAAA}
Sewon Min, Julian Michael, Hannaneh Hajishirzi, and Luke Zettlemoyer. 2020.
\newblock \href {https://doi.org/10.18653/v1/2020.emnlp-main.466} {{A}mbig{QA}:
  Answering ambiguous open-domain questions}.
\newblock In \emph{Proceedings of the 2020 Conference on Empirical Methods in
  Natural Language Processing (EMNLP)}, pages 5783--5797, Online. Association
  for Computational Linguistics.

\bibitem[{Naik et~al.(2018)Naik, Ravichander, Sadeh, Rose, and
  Neubig}]{Naik2018StressTE}
Aakanksha Naik, Abhilasha Ravichander, Norman Sadeh, Carolyn Rose, and Graham
  Neubig. 2018.
\newblock \href {https://www.aclweb.org/anthology/C18-1198} {Stress test
  evaluation for natural language inference}.
\newblock In \emph{Proceedings of the 27th International Conference on
  Computational Linguistics}, pages 2340--2353, Santa Fe, New Mexico, USA.
  Association for Computational Linguistics.

\bibitem[{Onoe and Durrett(2020)}]{Onoe2020FineGrainedET}
Yasumasa Onoe and Greg Durrett. 2020.
\newblock Fine-grained entity typing for domain independent entity linking.
\newblock In \emph{AAAI}.

\bibitem[{Orr et~al.(2020)Orr, Leszczynski, Arora, Wu, Guha, Ling, and
  R{\'e}}]{Orr2020BootlegCT}
Laurel Orr, Megan Leszczynski, Simran Arora, Sen Wu, Neel Guha, Xiao Ling, and
  Christopher R{\'e}. 2020.
\newblock \href {https://arxiv.org/abs/2010.10363} {Bootleg: Chasing the tail
  with self-supervised named entity disambiguation}.
\newblock \emph{arXiv preprint arXiv:2010.10363}.

\bibitem[{Petroni et~al.(2021)Petroni, Piktus, Fan, Lewis, Yazdani, De~Cao,
  Thorne, Jernite, Karpukhin, Maillard, Plachouras, Rockt{\"a}schel, and
  Riedel}]{Petroni2020KILTAB}
Fabio Petroni, Aleksandra Piktus, Angela Fan, Patrick Lewis, Majid Yazdani,
  Nicola De~Cao, James Thorne, Yacine Jernite, Vladimir Karpukhin, Jean
  Maillard, Vassilis Plachouras, Tim Rockt{\"a}schel, and Sebastian Riedel.
  2021.
\newblock \href {https://www.aclweb.org/anthology/2021.naacl-main.200} {{KILT}:
  a benchmark for knowledge intensive language tasks}.
\newblock In \emph{Proceedings of the 2021 Conference of the North American
  Chapter of the Association for Computational Linguistics: Human Language
  Technologies}, pages 2523--2544, Online. Association for Computational
  Linguistics.

\bibitem[{Prabhakaran et~al.(2019)Prabhakaran, Hutchinson, and
  Mitchell}]{Prabhakaran2019PerturbationSA}
Vinodkumar Prabhakaran, Ben Hutchinson, and Margaret Mitchell. 2019.
\newblock \href {https://doi.org/10.18653/v1/D19-1578} {Perturbation
  sensitivity analysis to detect unintended model biases}.
\newblock In \emph{Proceedings of the 2019 Conference on Empirical Methods in
  Natural Language Processing and the 9th International Joint Conference on
  Natural Language Processing (EMNLP-IJCNLP)}, pages 5740--5745, Hong Kong,
  China. Association for Computational Linguistics.

\bibitem[{Rajpurkar et~al.(2016)Rajpurkar, Zhang, Lopyrev, and
  Liang}]{Rajpurkar2016SQuAD10}
Pranav Rajpurkar, Jian Zhang, Konstantin Lopyrev, and Percy Liang. 2016.
\newblock \href {https://doi.org/10.18653/v1/D16-1264} {{SQ}u{AD}: 100,000+
  questions for machine comprehension of text}.
\newblock In \emph{Proceedings of the 2016 Conference on Empirical Methods in
  Natural Language Processing}, pages 2383--2392, Austin, Texas. Association
  for Computational Linguistics.

\bibitem[{Rao et~al.(2013)Rao, McNamee, and Dredze}]{Rao2013EntityLF}
Delip Rao, Paul McNamee, and Mark Dredze. 2013.
\newblock Entity linking: Finding extracted entities in a knowledge base.
\newblock In \emph{Multi-source, Multilingual Information Extraction and
  Summarization}.

\bibitem[{Ribeiro et~al.(2019)Ribeiro, Guestrin, and
  Singh}]{ribeiro-etal-2019-red}
Marco~Tulio Ribeiro, Carlos Guestrin, and Sameer Singh. 2019.
\newblock \href {https://doi.org/10.18653/v1/P19-1621} {Are red roses red?
  evaluating consistency of question-answering models}.
\newblock In \emph{Proceedings of the 57th Annual Meeting of the Association
  for Computational Linguistics}, pages 6174--6184, Florence, Italy.
  Association for Computational Linguistics.

\bibitem[{Ribeiro et~al.(2020)Ribeiro, Wu, Guestrin, and
  Singh}]{ribeiro-etal-2020-beyond}
Marco~Tulio Ribeiro, Tongshuang Wu, Carlos Guestrin, and Sameer Singh. 2020.
\newblock \href {https://doi.org/10.18653/v1/2020.acl-main.442} {Beyond
  accuracy: Behavioral testing of {NLP} models with {C}heck{L}ist}.
\newblock In \emph{Proceedings of the 58th Annual Meeting of the Association
  for Computational Linguistics}, pages 4902--4912, Online. Association for
  Computational Linguistics.

\bibitem[{Richardson et~al.(2020)Richardson, Hu, Moss, and
  Sabharwal}]{Richardson2020ProbingNL}
Kyle Richardson, H.~Hu, L.~Moss, and A.~Sabharwal. 2020.
\newblock Probing natural language inference models through semantic fragments.
\newblock In \emph{AAAI}.

\bibitem[{Sevgili et~al.(2020)Sevgili, Shelmanov, Arkhipov, Panchenko, and
  Biemann}]{Sevgili2020NeuralEL}
Ozge Sevgili, Artem Shelmanov, Mikhail~V. Arkhipov, Alexander Panchenko, and
  Christian Biemann. 2020.
\newblock \href {https://arxiv.org/abs/2006.00575} {Neural entity linking: A
  survey of models based on deep learning}.
\newblock \emph{arXiv preprint arXiv:2006.00575}.

\bibitem[{Shwartz et~al.(2020)Shwartz, Rudinger, and
  Tafjord}]{Shwartz2020YouAG}
Vered Shwartz, Rachel Rudinger, and Oyvind Tafjord. 2020.
\newblock \href {https://doi.org/10.18653/v1/2020.emnlp-main.556} {{``}you are
  grounded!{''}: Latent name artifacts in pre-trained language models}.
\newblock In \emph{Proceedings of the 2020 Conference on Empirical Methods in
  Natural Language Processing (EMNLP)}, pages 6850--6861, Online. Association
  for Computational Linguistics.

\bibitem[{Thorne et~al.(2018)Thorne, Vlachos, Christodoulopoulos, and
  Mittal}]{Thorne2018FEVERAL}
James Thorne, Andreas Vlachos, Christos Christodoulopoulos, and Arpit Mittal.
  2018.
\newblock \href {https://doi.org/10.18653/v1/N18-1074} {{FEVER}: a large-scale
  dataset for fact extraction and {VER}ification}.
\newblock In \emph{Proceedings of the 2018 Conference of the North {A}merican
  Chapter of the Association for Computational Linguistics: Human Language
  Technologies, Volume 1 (Long Papers)}, pages 809--819, New Orleans,
  Louisiana. Association for Computational Linguistics.

\bibitem[{Vrandecic and Kr{\"o}tzsch(2014)}]{Vrandecic2014WikidataAF}
Denny Vrandecic and M.~Kr{\"o}tzsch. 2014.
\newblock Wikidata: a free collaborative knowledgebase.
\newblock \emph{Commun. ACM}, 57:78--85.

\bibitem[{Warstadt et~al.(2020)Warstadt, Parrish, Liu, Mohananey, Peng, Wang,
  and Bowman}]{Warstadt2019BLiMPTB}
Alex Warstadt, Alicia Parrish, Haokun Liu, Anhad Mohananey, Wei Peng, Sheng-Fu
  Wang, and Samuel~R. Bowman. 2020.
\newblock \href {https://doi.org/10.1162/tacl_a_00321} {{BL}i{MP}: The
  benchmark of linguistic minimal pairs for {E}nglish}.
\newblock \emph{Transactions of the Association for Computational Linguistics},
  8:377--392.

\bibitem[{Wolf et~al.(2020)Wolf, Debut, Sanh, Chaumond, Delangue, Moi, Cistac,
  Rault, Louf, Funtowicz, Davison, Shleifer, von Platen, Ma, Jernite, Plu, Xu,
  Le~Scao, Gugger, Drame, Lhoest, and Rush}]{wolf-etal-2020-transformers}
Thomas Wolf, Lysandre Debut, Victor Sanh, Julien Chaumond, Clement Delangue,
  Anthony Moi, Pierric Cistac, Tim Rault, Remi Louf, Morgan Funtowicz, Joe
  Davison, Sam Shleifer, Patrick von Platen, Clara Ma, Yacine Jernite, Julien
  Plu, Canwen Xu, Teven Le~Scao, Sylvain Gugger, Mariama Drame, Quentin Lhoest,
  and Alexander Rush. 2020.
\newblock \href {https://doi.org/10.18653/v1/2020.emnlp-demos.6} {Transformers:
  State-of-the-art natural language processing}.
\newblock In \emph{Proceedings of the 2020 Conference on Empirical Methods in
  Natural Language Processing: System Demonstrations}, pages 38--45.
  Association for Computational Linguistics.

\bibitem[{Wu et~al.(2020)Wu, Petroni, Josifoski, Riedel, and
  Zettlemoyer}]{Wu2020ZeroshotEL}
Ledell~Yu Wu, F.~Petroni, Martin Josifoski, Sebastian Riedel, and Luke
  Zettlemoyer. 2020.
\newblock Zero-shot entity linking with dense entity retrieval.
\newblock In \emph{EMNLP}.

\bibitem[{Zhao et~al.(2018)Zhao, Wang, Yatskar, Ordonez, and
  Chang}]{Zhao2018GenderBI}
Jieyu Zhao, Tianlu Wang, Mark Yatskar, Vicente Ordonez, and Kai-Wei Chang.
  2018.
\newblock \href {https://doi.org/10.18653/v1/N18-2003} {Gender bias in
  coreference resolution: Evaluation and debiasing methods}.
\newblock In \emph{Proceedings of the 2018 Conference of the North {A}merican
  Chapter of the Association for Computational Linguistics: Human Language
  Technologies, Volume 2 (Short Papers)}, pages 15--20, New Orleans, Louisiana.
  Association for Computational Linguistics.

\end{thebibliography}
